\documentclass[10pt,journal]{IEEEtranTCOM}

\usepackage[english]{babel}
\usepackage[usenames]{color}
\usepackage[cp1250]{inputenc}
\usepackage{amsfonts}
\usepackage{amsthm}
\usepackage{graphicx}
\usepackage{epsfig}
\usepackage{mathrsfs}
\usepackage{amsmath}
\usepackage{algorithm}
\usepackage{algorithmic}
\usepackage{hyperref}
\usepackage{array}

\usepackage{etoolbox}
\makeatletter \patchcmd{\@makecaption}  {\scshape}  {}  {}  {}
\makeatother

\theoremstyle{plain}
\newtheorem{df}{Definition}

\newtheorem{thm}[df]{Theorem}

\begin{document}
\newcommand{\bea}{\begin{eqnarray}}
\newcommand{\eea}{\end{eqnarray}}
\newcommand{\be}{\begin{equation}}
\newcommand{\ee}{\end{equation}}
\newcommand{\beas}{\begin{eqnarray*}}
\newcommand{\eeas}{\end{eqnarray*}}
\newcommand{\bs}{\backslash}
\newcommand{\bc}{\begin{center}}
\newcommand{\ec}{\end{center}}
\def\SC {\mathscr{C}}
\newcommand\myeq{\mathrel{\stackrel{\makebox[0pt]{\mbox{\normalfont\tiny *}}}{=}}}

\title{Higher order PCA-like rotation-invariant features\\for detailed shape descriptors modulo rotation}
\author{\IEEEauthorblockN{Jarek Duda}, 
\IEEEauthorblockA{Jagiellonian University, Krakow, Poland,
Email: \emph{dudajar@gmail.com}}}
\maketitle

\begin{abstract}
PCA can be used for rotation invariant features, describing a shape with its $p_{ab}=E[(x_i-E[x_a])(x_b-E[x_b])]$ covariance matrix approximating shape by ellipsoid, allowing for rotation invariants like its traces of powers. However, real shapes are usually much more complicated, hence there is proposed its extension to e.g. $p_{abc}=E[(x_a-E[x_a])(x_b-E[x_b])(x_c-E[x_c])]$ order-3 or higher tensors describing central moments, or polynomial times Gaussian allowing decodable shape descriptors of arbitrarily high accuracy, and their analogous rotation invariants. Its practical applications could be rotation-invariant features to include shape modulo rotation e.g. for molecular shape descriptors, or for up to rotation object recognition in 2D images/3D scans maybe also for 3D scene understanding, or shape similarity metric allowing inexpensive comparison of objects modulo rotation avoiding costly optimization over rotations.
\end{abstract}
\section{Introduction}
In many tasks we work on freely shifting and rotating objects like molecules, optionally also scaling like digits - e.g. in image recognition searching and evaluating shapes in 2D dimensional images or 3D scans, molecules in chemoinformatics, etc. Rotation-invariant features are useful for such tasks, e.g. to include shape in $\mathbb{R}^d$ modulo $\textrm{SO}(d)$ rotation in models, inexpensively recognize rotated copies, or evaluate similarity between shapes of objects - without costly search through rotations.

For example in chemoinformatics there are used $d=3$ dimensional shape descriptors as distances from specific points~\cite{ultrafast}, or based on spherical harmonics~\cite{harm}, or fitting parabola and describing evolution through its cross-section~\cite{shape} - all providing description of shape having low details.

\begin{figure}[t!]
    \centering
        \includegraphics{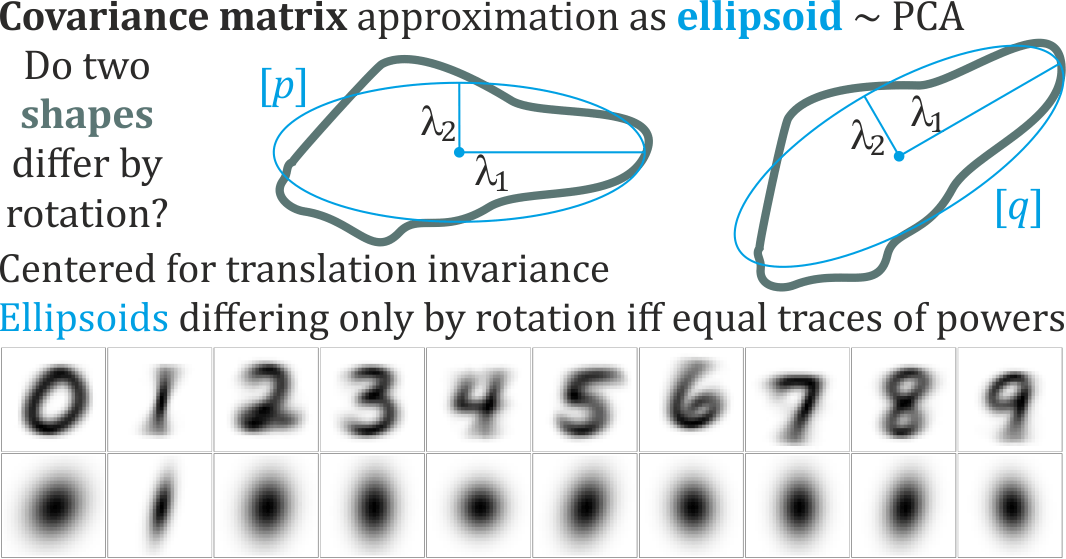}
        \caption{Having some shape/density, we can average ($E[]$) over it defining 2nd order polynomial with $[p]_{ab}\equiv p_{ab}=E[(x_a-E[x_a])(x_b-E[x_b])]$ covariance matrix, approximating this shape as ellipsoid. Finding such matrix for two shapes as $[p]$ and $[q]$, we can compare their eigenvalues/eigenvectors to find relative rotations/shapes with PCA-like approximation. To test differing by only rotation, we could check equality of $\forall_{i=1}^d \textrm{Tr}\left([p]^i\right) =\textrm{Tr}\left([q]^i\right)$ in $\mathbb{R}^d$ rotation invariants, agreement of all ensures similarity: $[p]\sim [q] \equiv \exists_{O\in \textrm{SO}(d)}: [p]=O[q]O^T$. As such ellipsoid approximation is insufficient to describe more complex shapes, here we propose to extend it to higher central moments, starting with $p_{abc}=E[(x_a-E[x_a])(x_b-E[x_b])(x_c-E[x_c])]$ order-3 tensor, for which we can extend $\textrm{Tr}([p]^i) =\textrm{Tr}([q]^i)$ rotation invariants with analogous graph-based invariants, starting with $\sum_{abc} (p_{abc})^2$. Bottom: averaged 28x28 MNIST~\cite{mnist} digits and such their PCA approximation.}
       \label{ex}
\end{figure}

Here we propose looking novel approach extending PCA(principal component analysis)-based approximation of shape as ellipsoid by covariance matrix (order-2) like in Fig. \ref{ex}, into higher order tensors e.g. central moments, or polynomials multiplied by Gaussian - as in Fig. \ref{Hermite} allowing as detailed shape description as needed, still allowing to decode this shape. Additionally it is continuous, preventing jumps in description for small shape changes, also allowing to describe dynamics e.g. by sets/sequences of invariants through molecular dynamics.

Related spherical harmonics go from description to invariants by sums of squares, this way losing information. Instead, the proposed descriptions by just polynomials allows for as many graph-based invariants as we want, hopefully allowing for complete description modulo rotation, however, finding complete sets of such invariant remains an open question.

This is early article proposing this looking novel family of methods, intended to be extended both from theoretical side, and its various applications.

\section{Representing shape/density as polynomial}
In this Section we discuss obtaining tensor/polynomial representation of given shape, originally being e.g. density/grayness from various scans or averaging, or volume defined by boundaries, or points of e.g. lattice or atoms of molecule. 

For multiple channels e.g. atomic properties or colors, we could e.g.  build separately such descriptions, also adding invariants mixing them to ensure common rotation.
\subsection{Shape represented as expected value function $f\to E[f(\mathbf{x})]$}
To represent shape as a polynomial, let us start with defining \emph{expected value} $E[f(\mathbf{x})]$ of various functions $f:\mathbb{R}^d\to \mathbb{R}$ for this shape, for example:
\begin{itemize}
  \item for density $\int \rho(\mathbf{x}) d\mathbf{x} =1$ it is $E[f(\mathbf{x})]=\int_{\mathbb{R}^d} f(\mathbf{x})\rho(\mathbf{x}) d\mathbf{x}$,
  \item for uniform volume $V$ it is $E[f(\mathbf{x})]=\int_V f(\mathbf{x}) d\mathbf{x}/|V|$,
  \item for weighted points: $E[f(\mathbf{x})]=\sum_\mathbf{x} \rho(\mathbf{x}) f(\mathbf{x}) d\mathbf{x}$ normalized $\sum_\mathbf{x}\rho(\mathbf{x}) =1$ - e.g. atoms of molecule, or lattice.
\end{itemize}

It allows to define e.g. the \emph{center of mass} $E[x_a]$, we can subtract from $x_a$ to normalize position - centering two shapes we would like to compare modulo translation.

\subsection{Covariance matrix $[p]$} 
Standard next step is defining $d\times d$ \emph{covariance matrix}:
\be [p]_{ab}\equiv p_{ab}=E\left[(x_a-E[x_a])(x_b-E[x_b])\right]\ee
allowing to approximate our shape with ellipsoid - of axes as $\textbf{v}^i$ eigenvectors for $[p]\textbf{v}^i=\lambda_i [p]$, and lengths $\sqrt{\lambda_i}$ from its eigenvalues. However, as for MNIST in Fig. \ref{ex}, such ellipsoid could  represent well only very simple shapes.

We can use such ellipsoid representation to orient rotation, e.g. rotating eigenvectors sorted by eigenvalues to canonical directions~\cite{shape}. However, it has potential \textbf{continuity problem} - slight change of eigenvalues can change their order, getting large differences of oriented shapes.

To avoid such continuity problem, we can directly compare rotation invariants instead, like \emph{similarity test}: check that $\textrm{Tr}([p]^i)=\textrm{Tr}([q]^i)$ for $i=1,\ldots,d$ - allowing to conclude that $[p]$ and $[q]$ differ only by rotations: that there exists $\textrm{SO}(d)$ orthogonal $O: OO^T = O^T O=I$, such that $[p]=O[q]O^T$.

\subsection{Optional scale normalization}
In image recognition e.g. letters have the same semantic meaning no matter the scale, hence we would like to also include scale invariance, possible by adding scale normalization.

To combine with test of differing by rotation, after centralization we should rescale all coordinates by the same value, e.g. dividing them by $x_i \to x_i/\sqrt{\textrm{Tr}([p])/d}$ we normalize to $\textrm{Tr}([p])=d$, making directional behavior on average approximately normalized Gaussian $N(0,1)$. It might be worth changing this $d$ to some different value to optimize representation.

\subsection{Central moments} The basic approach is just extending such order-2 covariance matrix into higher order-$r$ central moments with $r$ indexes:
\be p_{abc\ldots}=E\left[(x_a-E[x_a])(x_b-E[x_b])(x_c-E[x_c])\ldots\right]\label{cm}\ee

As permutation of indexes does not change its value, this is symmetric tensor - splitting $d$ dimensions into $r$ subsets of $\geq 0$ size, combinatorially getting
\be \textrm{number of }\mathbb{R}^d \textrm{ \textbf{symmetric} order-}r \textrm{ tensors: } {{d+r-1}\choose r} \label{sym}\ee

\subsection{Decodable representation: Gaussian times polynomial}
While we could work on the above symmetric tensors, they use abstract moments difficult to translate into the actual shape, and do not guarantee representing some unique shape.

If we need decodability of such description and representation of unique shape, we can approximate this density by a polynomial, what also brings completeness of representation:  polynomial approximations can be as close as needed, corresponding to some unique shape we could decode from it.

However, while shapes usually have finite size, polynomials go to infinity and explode - to represent shapes by polynomials, we should multiply them by some vanishing function, rather spherically symmetric for rotation invariance, like Gaussian $e^{-||\mathbf{x}||^2/2}$ for used $||\mathbf{x}||\equiv\sqrt{\mathbf{x}^T \mathbf{x}}$ Euclidean norm. 

\begin{figure}[t!]
    \centering
        \includegraphics[width=9cm]{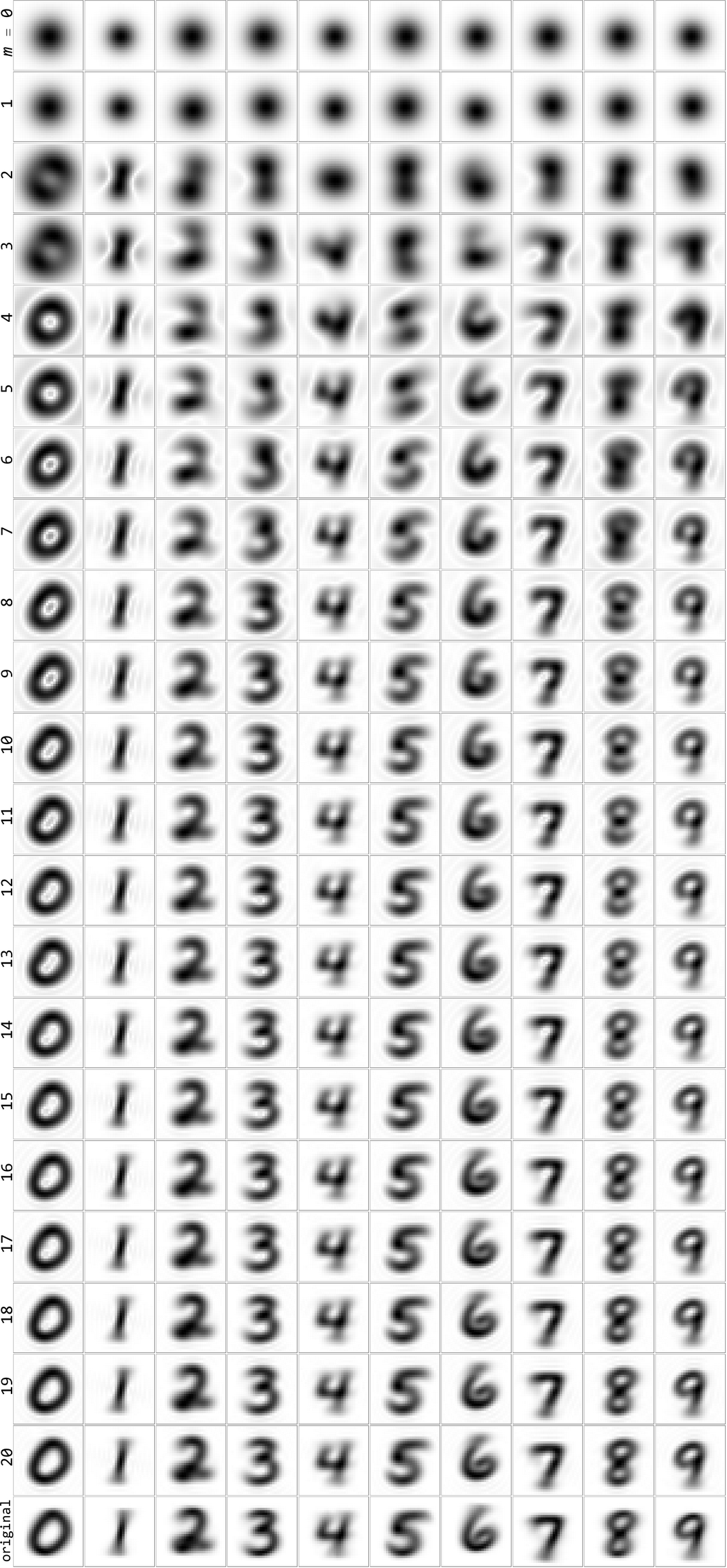}
        \caption{Visualization of accuracy of Gaussian times polynomial representation for MNIST~\cite{mnist} averaged digits as $d=2$ dimensional $28\times 28$ grayscale images, using $j_1+j_2=r$ for $r=0,..,m$ sums of polynomial degrees. Specifically, the grayness was normalized into density $\rho$ summing to 1 defining $E[]$, then center $(E[x_1], E[x_2])$ was subtracted from positions, then covariance matrix $[p]$ was calculated and coordinates were divided by $\sqrt{\textrm{Tr}([p])/d}$ for scale normalization. Then coordinates in product basis of (\ref{herm}) for $j_1+j_2\leq m$ were calculated, and used to reconstruct the images - which are shown. Due to replacing integration with summation over the lattice, the basis slightly loses orthonormality, what can be improved by Gram-Schmidt orthogonalization, at cost of less accurate representation. }
       \label{Hermite}
\end{figure}

Using orthonormal $\{f_i(x)\}$ basis: $\int f_i(x) f_j(x) dx = \delta_{ij}$, we can inexpensively MSE estimate density~\cite{rapid}:
\be \rho(x)=\sum_{j=0}^m u_j f_j(x) \quad \Rightarrow \quad u_j = \int \rho(x) f_j(x) dx = E[f_j(x)]\ee
Such orthonormal basis as polynomials times Gaussian is:
\be f_j(x) = \frac{1}{\sqrt{2^j j! \sqrt{\pi}}}\, h_j(x)\, e^{-x^2/2} \label{herm}\ee
where $h_i(x)$ are Hermite polynomials, for $i=0,\ldots,5$ being:
$$1,2x,4x^2-2,8x^3-12x,16x^4-48x^2+12,32x^2-160x^3+120x $$
In multidimensional situation we can use its product basis:
\be f_{\mathbf{j}}(\mathbf{x})\equiv f_{(j_1,..,j_d)}(x_1,..,x_d) = f_{j_1}(x_1)f_{j_2}(x_2)\cdots f_{j_d}(x_d)\ee 
To gain intuitions regarding accuracy of such representation, Figure \ref{Hermite} shows such $d=2$ representation for averaged MNIST digits, and $j_1+j_2\leq m$ for $m=0,\ldots,20$.


As applied $e^{-||\mathbf{x}||^2/2}$ weight has characteristic scale, this representation rather requires some scale normalization. If scale is not important (e.g. digits), we can normalize scale e.g. $x_i \to x_i /\sqrt{\textrm{Tr}([p])/d}$. If size needs to be distinguished, there should be chosen some fixed universal scaling.\\

Beside Gaussian, it might be also worth to consider different functions to multiply by polynomial, which for rotation invariance should depend only on radius $||\mathbf{x}||$ - e.g. zeroing outside some distance for \textbf{compact support}, or with \textbf{different power in exponent} (than 2) like in Laplace (1) or generally as in Exponential Power distribution~\cite{EPD}, or maybe of heavy tails 1/polynomial but restricting the highest finite moment.

Another approach to handle e.g. high anisotropy issue could be applying \textbf{radial rescaling}: $\mathbf{x} \to f(||\mathbf{x}||)\, \mathbf{x}/||\mathbf{x}||$ with e.g. $f(r)=r^p$ for some power $p$, then rescaled represent e.g. as Gaussian times polynomial.

We could also use anisotropic deformation like \textbf{whitening} applying $[p]^{-1/2}$ to all vectors first, transforming covariance matrix into identity $[p]\to I$, what is still continuous. Performing rotation earlier would not change found rotation invariants. However, this way we could not distinguish versions of objects with applied anisotropic rescaling - getting same invariants, what can be overcomed e.g. additionally including traces of powers of the original covariance matrix $[p]$ to used vector of invariants, maybe also mixed invariants (between original $[p]$ and further polynomial) to ensure applied same rotation.


\section{Rotation invariants for tensors/polynomials}
The proposed rotation invariants are calculated from tensors - as the above cental moments (\ref{cm}), or found polynomial coefficients (without Gaussian weight). Only the former are certain to be symmetric, their numbers are summarized in Table \ref{table}.

We can directly use these central moments, or split such polynomial $p:\mathbb{R}^d \to \mathbb{R}$ into fixed degree $r$ \textbf{homogenous} represented by \textbf{tensors} of this order $r$, up to chosen $m$ like in Fig. \ref{Hermite}, getting the polynomial representation we focus on:
\be p(\mathbf{x}) =\sum_{r=0}^m p^r(\mathbf{x})= p_\emptyset + \sum_{a=1}^d p_a x_a + \sum_{a,b=1}^d p_{ab} x_a x_b +\ldots \label{pol}\ee
\noindent where $p^r(\mathbf{x})$  is homogeneous degree $r$ polynomial, of scaling: 
\be p^r(\mathbf{x})=\|\mathbf{x}\|^r p^r(\hat{\mathbf{x}})\qquad \textrm{for}\quad \hat{\mathbf{x}}=\mathbf{x}/\|\mathbf{x}\|\ee

Due to symmetry, we can group indexes to $(x_1)^{j_1}\cdots (x_d)^{j_d}$, for  $j_1+\ldots+j_d=r$ numbers of appearances of each coordinate, having ${r \choose j_1,\ldots,j_d}$ copies in summations.\\

In \emph{tensor rank-$R$ decomposition}~\cite{tensor} generalizing SVD (singular value decomposition), we could express it as:
\be p^r = \sum_{i=1}^R \lambda_i\, v_{i,1}\otimes v_{i,2}\otimes \cdots \otimes v_{i,d}\label{dec}\ee

\subsection{Rotation invariance}
Having $p,q$ polynomials (\ref{pol}) describing two shapes, we would like to test if they differ only by (orthogonal) rotation:
\be p\sim q\qquad\equiv\qquad \exists_{O:O^TO=I}\ \quad p=q\circ O\ee
for $(q\circ O)$ rotating all $r$ inputs:
$$(q\circ O)(\mathbf{x})=q(O\mathbf{x})\quad\textrm{e.g.}\quad (q^3\circ O)_{ijk}=\sum_{abc} q_{abc} O_{ai} O_{bj} O_{ck}$$
For $r=0,1,2$ order tensors rotation invariants are well known:
\begin{itemize}
  \item $r=0$ it requires same value: $p_\emptyset = q_\emptyset$,
  \item $r=1$ requires same vector length: $\sum_{a=1}^d p_a^2 =\sum_{a=1}^d q_a^2$,
  \item $r=2$ requires similarity: $\forall_{i=1}^d\ \textrm{Tr}([p]^i)=\textrm{Tr}([q]^i)$.
\end{itemize}

One type of difficulty is making sure that $r=1$ and $r=2$ use the same rotation, what can be resolved by \textbf{mixed invariants}, like testing if $\forall_{i=1..d}\,  \sum_a p_a ([p]^i)_{ab}\, p_b = \sum_a q_a ([q]^i)_{ab}\, q_b$.

\begin{table}[t]
\begin{center}
\begin{tabular}{|c|c|c|c|c|c|c|c|}
  \hline
  $d$&rots&$r=1$&$r=2$&$r=3$&$r=4$&$r=5$&$r=6$ \\ \hline
  1 & 0 & 1/1 & 1/1 & 1/1 & 1/1 & 1/1 & 1/1\\
  \textbf{2} & 1 & 2/2 & 4/3 & 8/4 & 16/5 & 32/6 & 64/7 \\
  \textbf{3} & 3 & 3/3 & 9/6 & 27/10 & 81/15 & 243/21 & 729/28 \\
  4 & 6 & 4/4 & 16/10 & 64/20 & 256/35 & 1024/56 & 4096/84 \\
  5 & 10 & 5/5 & 25/15 & 125/35 & 625/70 & 3125/126 & 15625/210 \\
  \hline
\end{tabular}
\end{center}
\caption{\textbf{Dimensions of tensors}: \textbf{all}/\textbf{symmetric} using $d^r$ and ${d+r-1 \choose r}$ formulas for dimension $d=1..5$ and order $r=1..6$. Their complete description modulo rotation would require number of independent invariants lowered by the written dimension of rotations rots $=d(d-1)/2$. }
\label{table}
\end{table}

\begin{figure}[t!]
    \centering
        \includegraphics{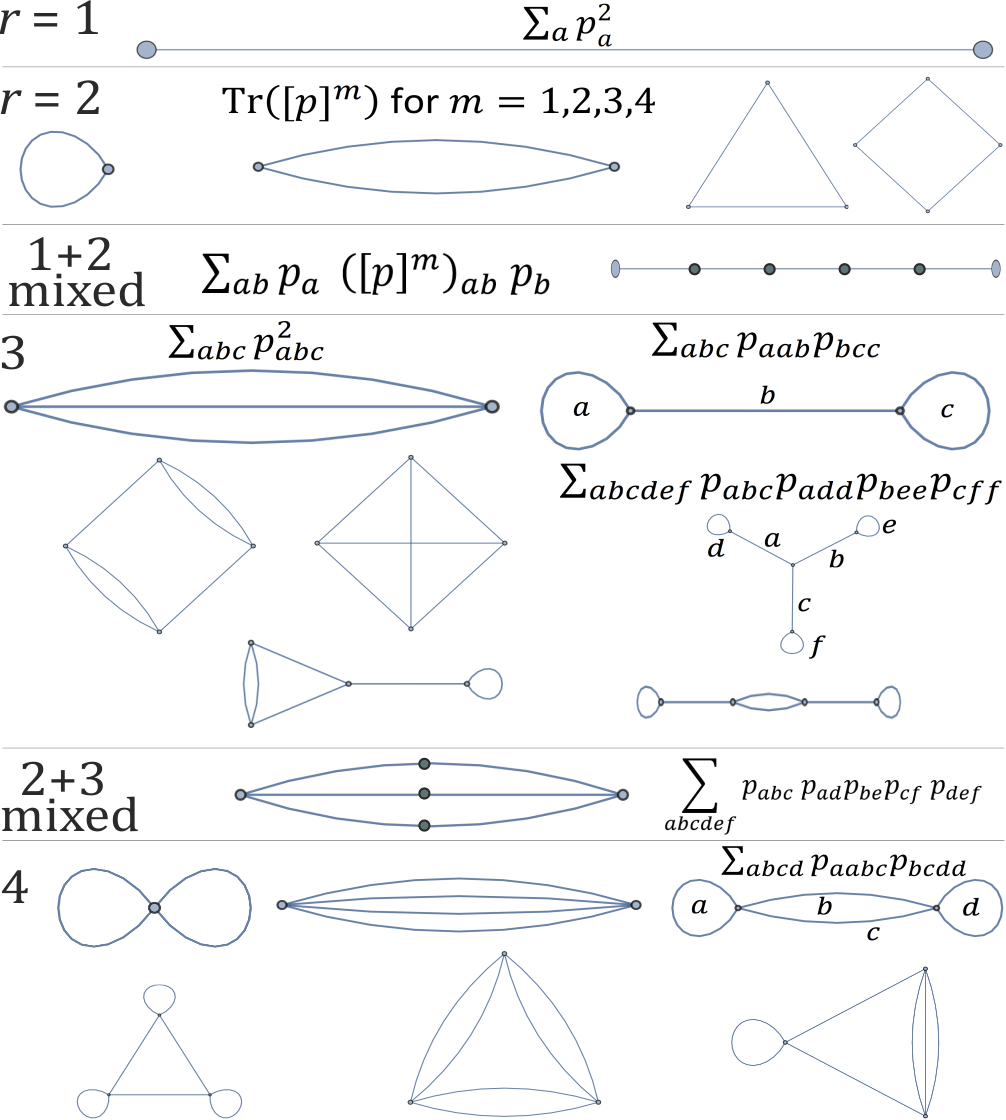}
        \caption{Diagrammatic representations~(\cite{pnp,poly}) of some first rotation invariants for degree 1, 2, 3, 4 homogeneous polynomials and mixed. Degree $r$ vertex corresponds to order-$r$ tensor e.g. in polynomial describing shape. For symmetric tensors edges for each vertex are indistinguishable. Every edge corresponds to summation over corresponding index like in matrix product, and is rotation invariant thanks to $\sum_i O_{ai}O_{b i}=\delta_{a b}$ relation for rotation $O$ applied to both indexes of given edge. Invariants from disconnected graphs can be omitted as being products over invariants for its components. Fig. \ref{aut} proposes systematic generation of larger numbers of such invariants. 
        For non-symmetric tensors we can use invariants for various permutations of tensor indexes. For descriptions of multiple properties e.g. colors, we can add invariants from mixed graphs having multiple types of vertices for these properties. For SO(1,3) Lorentz group instead of SO($d$), we can include $(-1,1,1,1)$ signature in products defined by edges.
        }
       \label{diag}
\end{figure}


\subsection{General rotation invariants}
\begin{figure}[t!]
    \centering
        \includegraphics{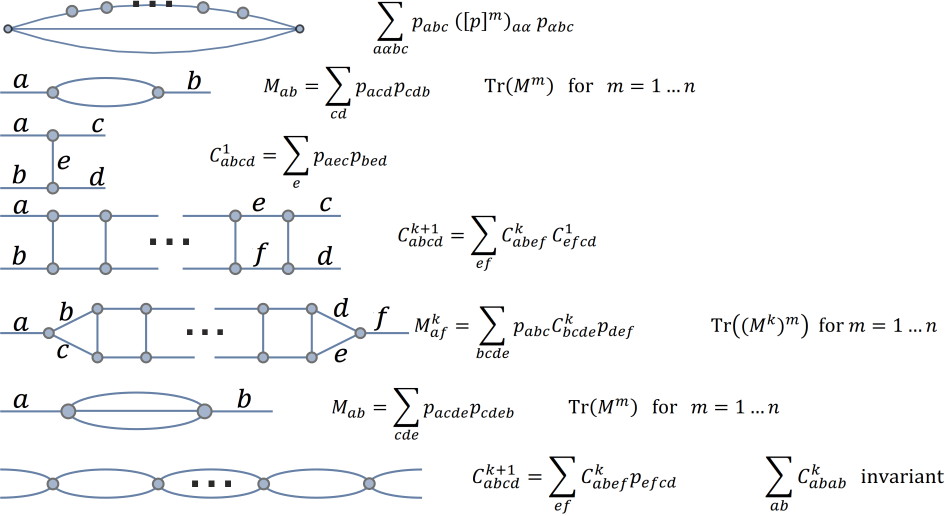}
        \caption{Some possibilities for systematic generation of large numbers of rotation invariants from \cite{pnp}, e.g. building $d\times d$ matrix $M$ (or $d^2 \times d^2$ matrix $C$) from order-3,4 tensors, then testing equality of $\textrm{Tr}(M^i)$ for $i=1,..,d$ (or $\textrm{Tr}(C^i)$ for $i=1,..,d^2$) for two tensors/polynomias/shapes, ensuring existence of orthogonal $d\times d$ (or $d^2 \times d^2$) between them. The dimension of $\textrm{SO}(d)$ rotations is $d(d-1)/2$, hence it should be sufficient to test this number of rotation invariants, however, mane of them are dependent like $\textrm{Tr}\left([p]^{i}\right)$ for $i=1,\ldots,d+1$ - the main difficulty is finding such number of independent invariants.}
       \label{aut}
\end{figure}

The above rotation invariants can be represented diagrammatically like in Fig. \ref{diag} ,\ref{aut}, also for higher orders - using graphs with vertices of degree $r$, corresponding to invariants by summing over dimensions $1,..,d$ for all edges. 

Applying any orthogonal matrix $OO^T=O^TO=I$ defining rotation, for each such edge it multiplies from one side by $O$, from the other by $O^T$, not changing the summation outcome - therefore, each such graph indeed defines rotation invariant.

For \textbf{non-symmetric} tensors we can use invariants for various index permutations. To ensure common rotation we can use \textbf{mixed invariants} for graphs of various degrees, or types of vertices e.g. for various channels. For e.g. $SO(1,3)$ \textbf{Lorentz group} we could include signature in products defined by edges.

However, while agreement of such invariants is \textbf{necessary conditions} for rotation invariance, to be certain that $[p]\sim[q]$ we would also need \textbf{sufficient condition}: a \textbf{complete set of invariants}, which agreement allows to conclude differing only by rotation. While for $r=1,2$ it is known, for higher orders it seems a difficult open problem, which resolution should also solve graph isomorphism problem~\cite{pnp}. We can easily find dimension of tensor modulo rotation e.g. in Table \ref{table}, giving required number of invariants, however, the difficult part is ensuring such number of independent among invariants.

The discussed applications are usually in $d=2,3$ having only 1 or 3 dimension of rotations $d(d-1)/2$. Using more invariants than required should allow for more robust redundant description - resistant to distortions. Found matchings can be further verified.


\textbf{Frobenius product} and norm naturally generalize to tensors as basic mixed invariants for two vertices $p, q$:  
\be \langle p^r,q^r\rangle = \sum_{a_1,..,a_r=1}^d p_{a_1..a_r}\ q_{a_1..a_r} \qquad\quad \|p^r\| =\sqrt{\langle p^r,p^r\rangle} \label{fr}\ee
\noindent which are analogously invariant to $p\to p\circ O,\ q\to q\circ O$. For various orders $r$ we can use summation, maybe $w_r$ weighted.

The discussed invariants are for $O(d)$: do not distinguish mirror versions, e.g. $A$ from $DAD$ for $D=\textrm{diag}(-1,1,1,\ldots,1)$, especially important for chemistry to distinguish enantiomers. Pfaffian for anti-symmetrized might allow to help with that for non-symmetric descriptions like polynomial times Gaussian, for 3D case naturally generalized to order-3 tensor, which anti-symmetrized in 3D has one parameter, which should be rotation invariant and change sign for mirror symmetry - allowing to distinguish enantiomers. If antisymmetrized order-3 tensor would be zero, we could analogously search for chiral asymmetry in tensor of order 6, 9, 12, or higher multiplicity of dimension.

\subsection{Similarity tests without symmetry}
For $d\times d$ matrices, symmetric have $d(d+1)/2$ dimension, but general have $d^2$. For complete description without rotation: symmetric need $d(d+1)/2-d(d-1)/2=d$ independent invariants fully described by basic similarity test (*) below. However, for general matrices we need $d^2-d(d-1)/2=d(d+1)/2$ independent invariants: additional $d(d+1)/2-d^2=d(d-1)/2$ above (*) basic invariants.

\href{https://en.wikipedia.org/wiki/Specht\%27s_theorem}{\textbf{Specht's theorem}}~\cite{specht} shows that for complex unitary case, by adding conjugate transpose and considering all (infinite number) words as sequences of products of $A$ and $A^*$:
\begin{thm}
  Two matrices $A$ and $B$ are unitarily equivalent if and only if $\operatorname{Tr} W(A, A^*) = \operatorname{Tr} W(B, B^*)$ for all words $W$.
\end{thm}
Later Pearcy~\cite{specht1} has improved it showing that equality for all words of at most $2d^2$ degree is sufficient to conclude unitary equivalence. It is also upper bound for basis size in real orthogonal case we are interested in here - especially in low $d=2,3$ dimension, where hopefully we should find practical complete sets of invariants also for at least low order tensors. 

\textbf{Basic similarity test} (*) gives $d$ rotation invariants:

\be (*)\quad \forall_{k=1..d}\,\mathrm{Tr}(A^k)=\mathrm{Tr}(B^k)\label{c1}\ee
ensures existence of orthogonal $O$ such that $A=OBO^T$ only if $A$ and $B$ are symmetric $d\times d$ matrices. 


For general nonsymmetric matrices it ensures equality of spectrum, but we have e.g. \href{https://en.wikipedia.org/wiki/Jordan_normal_form}{Jordan normal form} with blocks reducing numbers of eigenvectors. Dimension without rotations grows from $d$ for symmetric to $d(d+1)/2$ as in \textbf{upper-triangular matrix}, obtained e.g. with \href{https://en.wikipedia.org/wiki/Schur_decomposition}{Schur decomposition} by rotations.

In the discussed graph-based rotation invariants we can freely permutate indexes of matrices/tensors in vertices - without symmetry can include invariants for various index permutations. For matrices it means including transposed in cycles as in Specht's theorem. Some natural choice for $d(d+1)/2$ invariants is e.g.:
\be (**) \qquad\forall_{k=1}^d\, \forall_{l=0}^{k-1}\, \mathrm{Tr}\left(A^k (A^T)^l\right)=\mathrm{Tr}\left(B^k (B^T)^l\right)\label{c2}\ee

However, the difficult part is ensuring $d(d+1)/2$ independent among them - ideally there should be automatically chosen such complete set of invariants for which we are certain of independence, also for at least low order tensors.

Before finding automatic constructions, we can directly test independence e.g. by Jacobian criterion~\cite{jacobian} - calculate derivatives of all invariants over all variables, then rank of such matrix gives the number of independent invariants - at least locally showing dimension of variety described by these invariants, however, not necessarily globally. Mathematica can handle low dimensional cases this way, e.g. below code confirms including all $4^2 -6 =10$ independent rotation invariants for $d=4$ for general matrix, for $d=5$ assuming upper-triangular like after Schur decompostion: 

\begin{footnotesize}
\begin{verbatim}
M=Table[If[i>j,0,Subscript[a, Row[{i,j}]]],{i,d},{j,d}];
inv=Table[Tr[MatrixPower[M,k].MatrixPower[Transpose[M],l]]
     ,{k, d}, {l, 0, k - 1}];
MatrixRank[Table[D[Catenate[inv],v],{v,Variables[inv]}]]
\end{verbatim}
\end{footnotesize}

For higher could use Monte-Carlo: test this rank for multiple randomly chosen matrices (e.g. with automatic differentiation):
\begin{footnotesize}
\begin{verbatim}
Counts[Table[MatrixRank[jac/.Table[v -> RandomReal[]
, {v, Variables[jac]}]], 100]] 
\end{verbatim}
\end{footnotesize}

To better understand (\ref{c2}) $\mathrm{Tr}\left(A^k (A^T)^l\right)$ type invariants, we can find  $d\to d+1$ induction step for upper-triangular like in Schur decomposition, $a\in \mathbb{R}, v\in\mathbb{R}^d, A\in\mathbb{R}^{d\times d}$:

\be \operatorname{Tr}\left(\left(\begin{array}{cc} a & v^T \\  0 & A \\      \end{array} \right)^k 
\left(\begin{array}{cc} a & 0 \\  v & A^T \\      \end{array} \right)^l
 \right)=\ee
$$=\operatorname{Tr}\left(A^k (A^H)^l\right)+\sum_{i=0}^{k-1}\sum_{j=0}^{l-1} a^{k+l-i-j-2}\  v^H  A^i (A^H)^j v  $$

Finding complete set of invariants for tensors is even more challenging, but the basic approach is building larger matrices e.g $C_{ij,kl}=\sum_u p_{iku}\, p_{jlu}$ for H-shaped graph, and using $\textrm{Tr}(M^i)$ for $i=1..d^2$ as rotation invariants, maybe also Pfaffians generalized to tensors for anti-symmetrized. For higher order tensors we can e.g. similarly build even larger matrices, or use multiple edges like $M_{ij,kl}=\sum_{uv} p_{ikuv}\, p_{jluv}$, and so on. For non-symmetric tensor, like transposition for matrices, we can use such invariants with various permutations of tensor indexes.

\subsection{Including shape variability}
In practice such shapes often vary due to dynamics, e.g. of molecules. We could include it e.g. by replacing such single vector of rotation-invariant features, with their trajectory, set, or density e.g. as (multivariate) Gaussian (or HCR~\cite{HCR}) in feature space. Maybe also their Fourier analysis in time, auto-correlations (can be  multi-feature~\cite{MF}) e.g. to also include description of various vibrations, their frequencies.

For example by performing molecular dynamics of given molecule, and regularly calculating invariants of snapshots, maybe finally estimating their distribution in space of such vectors as e.g. multivariate Gaussian. Further it could be directly compared with shape and dynamics of target protein binding site.

\section{Rotation optimization, shape similarity metric}
While equality of a complete set of rotation invariants would require differing only by rotation, in practice there are usually also deformations.

Hence we should have some distance evaluating such distortion, e.g. Frobenius (\ref{fr}) just summing squares of all coefficients of matrix/tensor/polynomial, and minimize it over rotations:
\be \min_{O:OO^T=I} \|p - q\circ O\|\qquad \textrm{for}\quad (q\circ O)(\mathbf{x})=q(O\mathbf{x}) \label{sim} \ee

While it resembles orthogonal Procrustes problem~\cite{proc}, it assumed rotation of only a single coordinate, what can be extended to order-$r$ tensor by just treating it as $d\times d^{r-1}$ matrix.

However, here we rotate all coordinates with the same $O$, making it much more difficult, resembling diagonalization problem: $ \max_{O:OO^T=I} \sum_a ((OMO^T)_{aa})^2 =\sum_a \lambda_a^2$. For order-$r$ tensor dimension grows $O(d^r)$, while it is $O(d^2)$ for rotations ($d(d-1)/2$), generally no longer allowing for diagonalization.

In practice optimization like (\ref{sim}) would rather require e.g. gradient descent method, likely having multiple local minima.\\

To \textbf{avoid this costly optimization problem}, rotation-invariant features allow to calculate vectors describing shape modulo rotation, and use some distance between them as evaluation of shape similarity, becoming 0 if they differ only by rotation.

For example for order-2 $[p]$ approximating shape as ellipsoid, discussed standard rotation invariants can be written as $d$ dimensional vector, for example with applied some order root to make them closer to generalized means of eigenvalues:
\be \textrm{inv}([p]) =\left(\left(\textrm{Tr}([p]^i)\right)^{1/i}\right)_{i=1..d}= \left(\sqrt[i]{\sum_a (\lambda_a)^i}\right)_{i=1..d}  \ee

We can analogously add more discussed rotation-invariant features to such vectors, e.g. like in \ref{diag}, \ref{aut}, using roots of e.g. order as the number of vertices of applied graph. Then evaluate similarity between two shapes (modulo rotation) by some distance between two such vectors of features $d(\textrm{inv}([p]),\textrm{inv}([q]))$.
\subsection{Shape similarity metric for matrices}
While finally we would like to include tensors, let us start with matrix case, e.g. symmetric covariance for basic PCA approach:

\be \min_{O:OO^T=I} \| A -OBO^T\| \approx d(\textrm{inv}(A),\textrm{inv}(B)) \label{mnorm}\ee

We can find necessary condition of zeroing derives by $\epsilon\equiv \epsilon_{kl}$ step in infinitesimal antisymmetric transformation for $(E_{kl})_{ij}=\delta_{ki}\delta_{lj}$ being $d\times d$ matrix with single 1 in some $k\neq l$ position:

$$B\to \left(I+\epsilon (E_{kl}-E_{lk})\right) B \left(I-\epsilon (E_{kl}-E_{lk})\right)$$
$$ B_{ij}\to B_{ij} +\epsilon\left(\delta_{ik} B_{lj}-\delta_{il} B_{kj}-B_{ik}\delta_{lj}+B_{il}\delta_{kj}
\right)+O(\epsilon^2)
$$
$$ \partial_\epsilon \| A -B\|=\sum_{ij} (\partial_{ij} \| A -B\|)\, \partial_\epsilon B_{ij}$$
we could also use for gradient optimization, what should be computationally much less expensive for matrix/tensor/polynomial representing shape, than directly working on this shape.

Assuming Frobenius norm, symmetric $A$, $B$, and diagonalized $A$, we get necessary condition that $B$ has to be also diagonalized, making norm (\ref{mnorm}) sum of squared differences between sorted eigenvalues of $A$ and $B$. However, using discussed traces of powers invariants instead, they are sums of powers of eigenvalues, e.g.  \href{https://en.wikipedia.org/wiki/Newton\%27s_identities}{Newton's identities} allows to translate to different symmetric polynomials of eigenvalues.

In $d=2$ formula translating invariants: $p_k=(\sum_{i=1}^2 (\lambda_i)^k)^{1/k}$, $q_k=(\sum_{i=1}^2 (\eta_i)^k)^{1/k}$, and such Frobenius norm for sorted eigenvalues looks reasonable:
\be \sum_{i=1}^2 (\lambda_i-\eta_i)^2= -\sqrt{2 p_2^2-p_1^2} \sqrt{2 q_2^2-q_1^2}-p_1 q_1+p_2^2+q_2^2 \ee

However, in higher dimension it becomes much complicated, and in practice such norms should be even more complex like discussed in the next Subsection, matrices do not need to be symmetric, and we would like to extend to tensors - making such distances between vectors of invariants even more complicated, in practice could be e.g. approximated as trained neural networks.

\subsection{Standard density distances}
Training such distance between vectors of invariants as e.g. neural networks, we could use more appropriate distance between shapes, like standard e.g. Hausdorff or Tanimoto~\cite{tanimoto}.

Here we rather need to work on densities instead, standard choices are e.g. MSE, Kullback-Leibler, Wasserstein, or based on CDF like Kolmogotov-Smirnov test. The first two have closed formulas for multivariate Gaussians - below using $\Sigma$, $\Gamma$ covariance matrices shifted by vector $\mu$. Denoting $|M|=\det(M)$:
\be\rho_{\mu, \Sigma}(\mathbf{x}):= \frac{1}{\sqrt{|2\pi\Sigma|}}
e^{-\frac 12 (\mathbf{x}-\mu)^T\Sigma^{-1} (\mathbf{x}-\mu)}\ee
\subsubsection{Mean squared error (MSE)} is the basic density distance. Integral of product of Gaussians can be calculated~\cite{GAE}:
$$\int_{\mathbb{R}^d} \rho_{\mu, \Sigma}(\mathbf{x})\cdot\rho_{\mathbf{0},\Gamma}(\mathbf{x})\,d\mathbf{x}=\frac{\exp\left(-\frac 12 (\mu^T\Sigma^{-1}(\Sigma^{-1}+\Gamma^{-1})^{-1} \Gamma^{-1}\mu) \right)}
{\sqrt{(2\pi)^d |\Sigma||\Gamma||\Sigma^{-1}+\Gamma^{-1}|}}$$
for $\mu=0$ MSE as $\int_{\mathbb{R}^d} (\rho_{\mathbf{0},\Sigma}(\mathbf{x})-\rho_{\mathbf{0},\Gamma}(\mathbf{x}))^2 d\mathbf{x}$ becomes:
\be \sqrt{(2\pi)^d}\ \|\Sigma,\Gamma\|_{MSE}=\frac{1}{2|\Sigma|}+\frac{1}{2|\Gamma|}- \frac{1}
{|\Sigma||\Gamma||\Sigma^{-1}+\Gamma^{-1}|}\label{MSEd}\ee

\subsubsection{Kullback-Leibler divergence} is a basic choice for probability densities, and \href{https://stats.stackexchange.com/questions/60680/kl-divergence-between-two-multivariate-gaussians}{analytical formula for Gaussians can be found}:
$$D_{KL}=\frac{1}{2}\left( \operatorname{Tr}(\Gamma^{-1} \Sigma)+\mu^T \Gamma^{-1} \mu-d +
\ln\left( \frac{|\Gamma|}{|\Sigma|}\right)
\right)$$
taking $\mu=0$ and symmetrizing to Jensen-Shannon divergence:
\be \|\Sigma,\Gamma\|_{JS}   =\frac{1}{2}\left(
\operatorname{Tr}(\Gamma^{-1} \Sigma) + 
\operatorname{Tr}(\Sigma^{-1} \Gamma)\right)-d \label{JSd}
\ee
\subsubsection{Wasserstein/earth mover distance} is another basic choice, however, difficult and costly to calculate. For Gaussians there is known below upper bound \cite{earth} (plus $\|\mu\|^2$ for nonzero) we could use as approximation:
\be \|\Sigma,\Gamma\|_{W}\leq \operatorname{Tr}\left(\Sigma+\Gamma+2\sqrt{\Sigma \Gamma}\right) \label{Wd}\ee

\section{Conclusion and further work}
There was presented general approach to express shapes with polynomial (e.g. multiplied by Gaussian) by tensors, for which we can inexpensively calculate invariants of shifts, rotations and optionally scale - offering \textbf{detailed continuous decodable shape description}. Working on vectors of such invariants we can inexpensively e.g. include molecule shape modulo rotation for drug design, find rotated similar shapes, or estimate shape similarity avoiding costly rotation optimization.

This is early article proposing such looking novel approach, leaving many open questions both theoretical and practical, e.g.:

\begin{itemize}
  \item Search for complete set of invariants for order 3 and higher - which agreement ensures differing only by rotation.
  \item Designing shape similarity metrics based on such invariants, e.g. as some distance between vectors of chosen subset of invariants - allowing to inexpensively evaluate difference between two shapes modulo rotation.
  \item Optimization for various applications, like molecular shape descriptions, 2/3D image recognition, shape comparison.  
  \item Including shape variability crucial for various 2/3D objects e.g. molecules, maybe together with of charge distribution, or of binding site. Maybe with Fourier/autocorrelation to include description of vibrations.
  \item Maybe applications for 3D scene understanding for vision, graphics, robotics - searching database of objects remembered by rotation invariants, e.g. based on 2D projections.
  \item While scenes are usually built of triangles or Gaussians, it might be worth to consider more sophisticated objects like Gaussian times polynomial.
  \item While we have focused on $O(d)$ rotation invariants, in physics there are popular for $SO(1,3)$~\cite{complete} Lorentz group, we could revisit with discussed graph-based invariants - including e.g. $(-1,1,1,1)$ signature in sums over index represented by edges. 
\end{itemize}

\bibliographystyle{IEEEtran}
\bibliography{cites}
\end{document}